\documentclass[preprint,12pt]{elsarticle}

\usepackage{microtype}
\usepackage{graphicx}
\usepackage{booktabs}

\usepackage{amsmath}
\usepackage{amssymb}
\usepackage{mathtools}
\usepackage{amsthm}
\usepackage{enumitem}

\usepackage{tikz}
\usepackage{pgfplots}
\usepackage{subcaption}
\usetikzlibrary{arrows.meta, positioning, calc}

\usepackage[colorlinks=true,linkcolor=blue,citecolor=blue,urlcolor=blue]{hyperref}

\begin{document}
\date{}
\begin{frontmatter}

% -------- Title --------
\title{Energy‑Efficient Deep Reinforcement Learning with Spiking Transformers}

\author[uwyo]{Mohammad~Irfan~Uddin}
\author[uwyo]{Nishad~Tasnim}
\author[uwyo]{Md~Omor~Faruk}
\author[uwyo]{Zejian~Zhou\corref{cor1}}
% \ead{zzhou2@uwyo.edu}

\address[uwyo]{Department of Electrical Engineering and Computer Science, University of Wyoming, Laramie, WY~82071, USA}

% \cortext[cor1]{Corresponding author}

% -------- Abstract --------
\begin{abstract}
Agent‑based Transformers have been widely adopted in recent reinforcement learning advances due to their demonstrated ability to solve complex tasks. However, the high computational complexity of Transformers often results in significant energy consumption, limiting their deployment in real‑world autonomous systems. Spiking neural networks (SNNs), with their biologically inspired structure, offer an energy‑efficient alternative for machine learning. In this paper, a novel Spike‑Transformer Reinforcement Learning (STRL) algorithm that combines the energy efficiency of SNNs with the powerful decision‑making capabilities of reinforcement learning is developed. Specifically, an SNN using multi‑step Leaky Integrate‑and‑Fire (LIF) neurons and attention mechanisms capable of processing spatio‑temporal patterns over multiple time steps is designed. The architecture is further enhanced with state, action, and reward encodings to create a Transformer‑like structure optimized for reinforcement learning tasks. Comprehensive numerical experiments conducted on state‑of‑the‑art benchmarks demonstrate that the proposed SNN Transformer achieves significantly improved policy performance compared to conventional agent‑based Transformers. With both enhanced energy efficiency and policy optimality, this work highlights a promising direction for deploying bio‑inspired, low‑cost machine learning models in complex real‑world decision‑making scenarios.
\end{abstract}

% -------- Keywords --------
\begin{keyword}
Spiking Neural Networks \sep Transformer \sep Reinforcement Learning 
\end{keyword}

\end{frontmatter}

\section{Introduction}
\label{sec:intro}
Deep reinforcement learning (DRL) has proven to be a powerful tool for solving sequential decision-making tasks such as robotic control\citep{JIANG2021389,LI201992}, autonomous driving\citep{ZHOU2025128669}, game playing\citep{vinyals2019alpha}, and resource management\citep{YAN2022107688}. However, traditional DRL methods often struggle with long-term tasks due to the limited ability of regular neural networks (NNs) to capture extended temporal structures. Some efforts have been made to address this challenges. For example, previous literature has explored the use of LSTMs\citep{10.1162/neco.1997.9.8.1735} and RNN\citep{ELMAN1990179} architectures, which showed moderate improvements. More recently, transformers have emerged as state-of-the-art models for sequence modeling in domains such as natural language processing \citep{NIPS2017_3f5ee243,DBLP:conf/naacl/DevlinCLT19} and computer vision \citep{Dosovitskiy2020AnII}, due to their ability to learn long-range dependencies and capture complex patterns. However, transformers typically rely on very large scale computing intensive unit arrays (GPUs), which limits their direct applicability to real-world physical systems which usually has energy constraints. One of the most notable pioneers of applying transformers in DRL is the Decision Transformer algorithm \citep{chen2021decisiontransformerreinforcementlearning}. Despite the transformer’s proven ability to handle longer-horizon tasks, both training and inference require significant computational resources, such as GPUs—resources often unavailable in real-world autonomous systems that rely on DRL for long-term control. In this paper, we propose using an energy-efficient, bio-inspired neural network, i.e, a spiking neural network (SNN), to reconstruct the transformer architecture for DRL applications.

The third generation bio-inspired neural network, i.e., \emph{spiking neural networks} (SNNs), has gained increasing prominence for energy-efficient computation \citep{neftci2019surrogategradientlearningspiking}. By emulating the event-driven spiking mechanism found in biological neurons, SNNs can process information sparsely over time, potentially reducing computational overhead and aligning with specialized neuromorphic hardware \citep{8259423, doi:10.1126/science.1254642}. Empirical studies confirm that this sparsity translates into markedly lower energy use than conventional ANNs: inference on Intel’s \emph{Loihi} chip requires \(10\times\!–\!100\times\) less energy per image than a GPU/CPU running an equivalent DNN \citep{10.1145/3320288.3320304}. IBM’s \emph{TrueNorth} achieves an average \emph{46\,pJ} per synaptic operation—around two orders of magnitude more efficient than state-of-the-art CMOS accelerators \citep{doi:10.1073/pnas.1604850113}. Complementary results on \emph{SpiNNaker2} show up to a \(\!20\times\) reduction in joules-per-inference for spiking ResNets compared with quantized ANN counterparts executed on embedded ARM cores \citep{10.3389/fnins.2022.1018006}. These benchmarks collectively demonstrate the promise of SNNs for energy-constrained sequential decision-making systems. Despite these advantages, scaling SNNs to handle complex, long-horizon tasks remains an open challenge. While SNNs inherently capture temporal features, many existing architectures struggle to effectively model extended temporal and spatial dependencies required for RL tasks that demand sophisticated foresight and planning \citep{10.3389/fnins.2022.877701, tang2020reinforcementcolearningdeepspiking}. Efforts have been made to enhance temporal extraction.For example, surrogate-gradient frameworks such as SLAYER propagate errors across hundreds of simulation steps and improve speech and gesture recognition \citep{shrestha2018slayerspikelayererror}; the e-prop algorithm introduces local eligibility traces that approximate back-propagation-through-time in recurrent SNNs, enabling learning over thousands of time steps \citep{bellec2020solution}; spatio-temporal backpropagation with explicit timing-dependent objectives extends the effective temporal receptive field of convolutional SNNs \citep{10.3389/fnins.2018.00331}; and residual SNNs with learnable membrane constants deepen temporal integration without vanishing spikes \citep{fang2021incorporatinglearnablemembranetime}. However, a recent research shows that transformer-based approach using batch normalization strategies and novel attention modules can significantly boost accuracy on SNN's long horizon tasks. \citep{GAO2024128268}. 

\begin{figure}               % [t] = top of column; use [h] or [b] if preferred
  \centering
  \vspace{-6em} 
  \includegraphics[width=0.50\columnwidth]{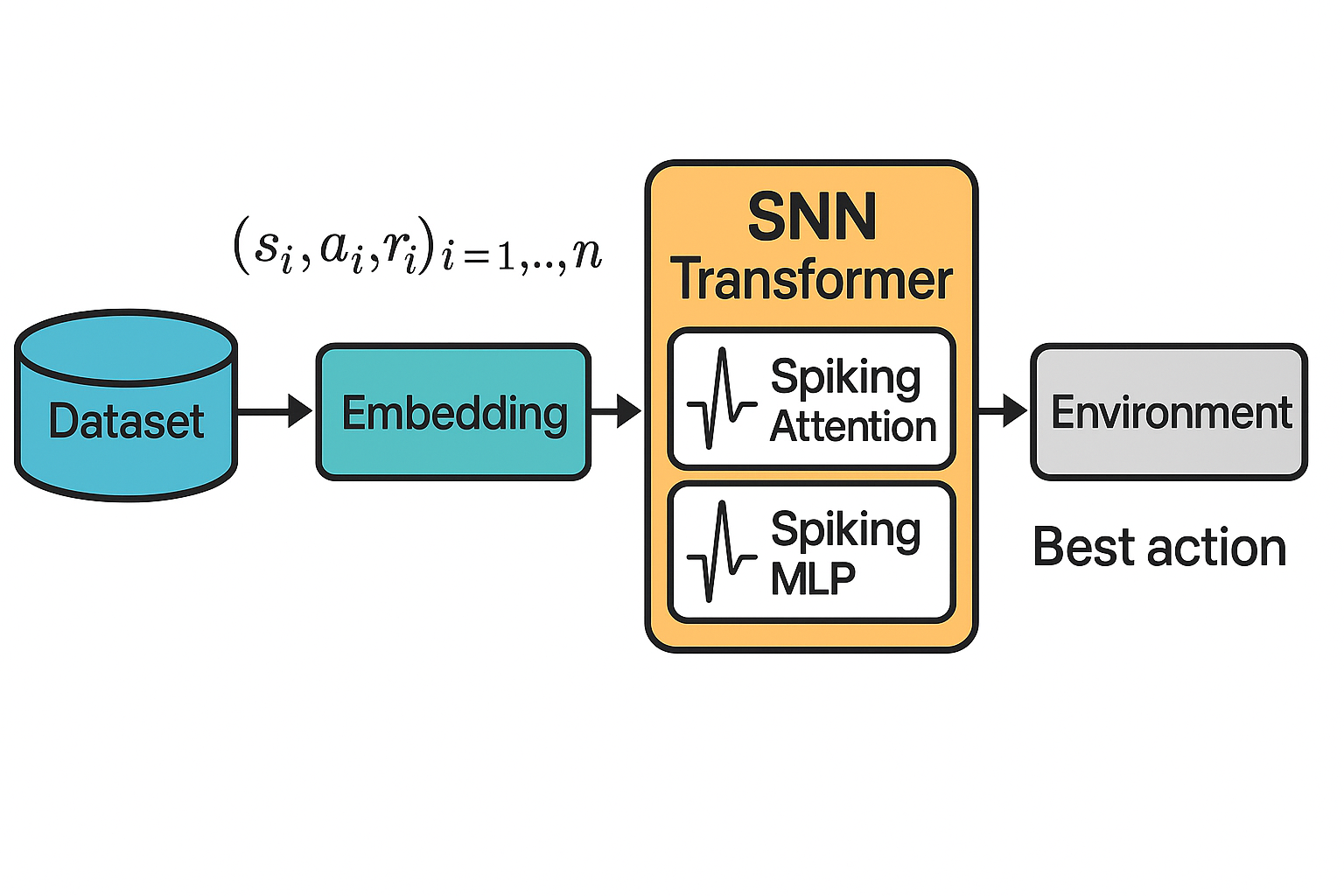}
  \vspace{-2em}
  \caption{ SNN-Transformer Overview.}
  \label{fig:gra}
  % \vspace{-1em}
\end{figure}

In this paper, a novel Spike Transformer-based reinforcement learning algorithm is designed. This work brings together the complementary strengths of Transformers and spiking neural networks in a single architecture to tackle offline sequential decision-making using deep reinforcement learning. Specifically, a \textbf{SNN Transformer} (Figure~\ref{fig:gra}) that integrates multi-head self-attention with multi-step Leaky Integrate-and-Fire (LIF) \citep{gerstner2002spiking} neurons is proposed. This design encodes state, action, and return-to-go embeddings, leveraging the transformer’s capacity for sequence modeling while introducing the sparse and biologically plausible dynamics of spiking neurons. The framework effectively addresses the temporal credit assignment problem in RL by modeling extended trajectories—an essential trait for tasks like maze navigation, where early decisions can drastically influence long-term outcomes.

The main contributions of the paper can be summarized as:
\begin{itemize}[itemsep=0.4em, topsep=0em, partopsep=0pt, parsep=0pt, leftmargin=*]
    \item A \emph{SNN Transformer} architecture for Sequential RL tasks is introduced to augment standard Transformers with multi-step LIF neurons, thereby providing both long-range attention capabilities and sparse spiking dynamics.
    \item The proposed SNN Transformer algorithm is energy-efficient due to the sparse and event-driven nature of spiking neural networks, which significantly reduces unnecessary computations. Additionally, by leveraging the temporal dynamics of spikes, it avoids continuous activation updates typical in standard Transformers, leading to lower power consumption during inference and training.
    % \item Demonstrate the efficacy of the proposed model on two substantial maze-navigation benchmarks (procedural mazes and D4RL), achieving near-perfect training accuracy and robust test performance.
    \item Demonstrate the effectiveness of the proposed SNN Transformer on a large-scale maze-navigation benchmark, achieving near-perfect test accuracy (over \textbf{99\%}) and robust generalization. 
\end{itemize}

\section{Background}
\label{sec:related_work}

Transformer‑based architectures have become increasingly popular in reinforcement learning (RL) owing to their capacity to handle long‑range temporal dependencies.  In particular, the \emph{Decision Transformer} (DT) \citep{chen2021decisiontransformerreinforcementlearning} pioneered casting trajectories as language sequences, where a Transformer predicts actions conditioned on states and returns‑to‑go.  Subsequent work has explored trajectory stitching for improved sample efficiency \citep{janner2021offlinereinforcementlearningbig}, multi‑task variants that share a single policy across heterogeneous domains \citep{Ghanem2023}, and hierarchical extensions that introduce options or sub‑policies to better capture temporal abstraction \citep{Correia2023}.  Recent studies also integrate contrastive pre‑training \citep{pmlr-v205-konan23a}, incorporate uncertainty estimation into the return‑to‑go token \citep{li2024uncertaintyawaredecisiontransformerstochastic}, and apply DTs to real‑world robotic manipulation \citep{brohan2023rt1roboticstransformerrealworld}.  Despite these advances, the core computational graph remains a stack of dense self‑attention and feed‑forward blocks executed at every time step, leading to millions of floating‑point multiply–accumulate operations per trajectory.  While these models effectively learn policies from diverse offline datasets, they continue to rely on dense floating-point operations, rendering them power-intensive for certain hardware deployments.

\section{Problem Formulation}
\label{sec:problem_formulation}
This paper investigates an episodic decision-making problem within the framework of a Markov Decision Process (MDP), where an agent interacts with an environment to achieve a designated goal. Formally, an MDP is defined by a state space \(\mathcal{S}\), an action space \(\mathcal{A}\), transition dynamics \(P(s' \mid s, a)\), and a reward function \(R(s, a)\). At each timestep \(t\), the agent observes its current state \(\mathbf{s}_t \in \mathcal{S}\) and selects an action \(a_t \in \mathcal{A}\) based on a policy \(\pi(a_t \mid \mathbf{s}_{1:t}, a_{1:t-1}, G_{1:t}, t_{1:t})\). The environment then transitions to a new state \(\mathbf{s}_{t+1}\) according to the transition probabilities \(P(s' \mid s, a)\) and provides a reward \(r_t\) based on the reward function \(R(s, a)\).

% As a specific case study, we apply this MDP framework to a discrete 2D maze environment. In this scenario, the state \(\mathbf{s}_t\) represents the agent's coordinates within a grid of size \(W \times H\), where each cell is either ``free'' (0) or ``blocked'' (1). The action space \(\mathcal{A} = \{\text{left}, \text{right}, \text{up}, \text{down}\}\) consists of four discrete movements that the agent can perform to navigate the maze. The objective is for the agent to determine an optimal sequence of actions that leads it from a starting position to a goal state, maximizing cumulative rewards while minimizing traversal costs.

% A trajectory \(\tau = (\mathbf{s}_1, a_1, r_1, \dots, \mathbf{s}_T, a_T, r_T)\) unfolds over at most \(T_{\max}\) time steps, with reward signal:
% \begin{itemize}
%     \item A small negative penalty (\(-0.1\)) per step to encourage shorter paths.
%     \item A positive terminal reward (\(+1.0\)) if the agent reaches the goal.
% \end{itemize}
The objective is to solve an optimal policy $\pi: \mathcal{S}\rightarrow \mathcal{A} $ that maximizes the expected return \citep{Mnih2015HumanlevelCT, 712192}
\begin{equation}
J(\pi_\theta) \;=\; \mathbb{E}_{\tau \sim \pi_\theta} \Biggl[\sum_{t=1}^T r_t \Biggr]
\end{equation} In DRL \citep{Mnih2015HumanlevelCT}, the policy is approximated by a deep neural network \(\pi_\theta\) parameterized by \(\theta\). However, the regular neural networks struggle in sequential information decoding, especially in long-horizon tasks. 

% In practice, an \emph{offline} RL scenario is adopted by leveraging a dataset of expert demonstrations \(\mathcal{D} = \{\tau^{(i)}\}_{i=1}^{N}\), where each \(\tau^{(i)}\) is an \emph{optimal} or near-optimal trajectory in a procedurally generated maze of size \(21 \times 21\). These demonstrations are obtained via A* search, which provides state-action pairs and associated step rewards. We then train a sequence model \(f_\theta\) to predict the next action \(a_t\) given \(\mathbf{s}_t\) (the current state), \(\{a_1,\dots,a_{t-1}\}\) (the past actions), and \(\{G_1,\dots,G_t\}\) (the returns-to-go), where
% \begin{equation}
% G_t \;=\; \sum_{\tau=t}^{T} r_\tau.
% \end{equation}
% This approach is akin to \emph{behavior cloning} or supervised trajectory learning, but it retains key RL concepts by including returns-to-go in the model input. The final policy emerges from minimizing cross-entropy loss on the expert actions under various maze configurations.

\section{Spike-Transformer Reinforcement Learning}
\label{sec:approach}
State-of-the-art DRL algorithms often struggle to effectively interpret sequential information. To overcome this limitation, the Spike-Transformer Reinforcement Learning (STRL) algorithm is proposed in this section, aiming to combine the energy-efficient temporal dynamics of spiking neural networks with the long-horizon decision-making capabilities of Transformers.

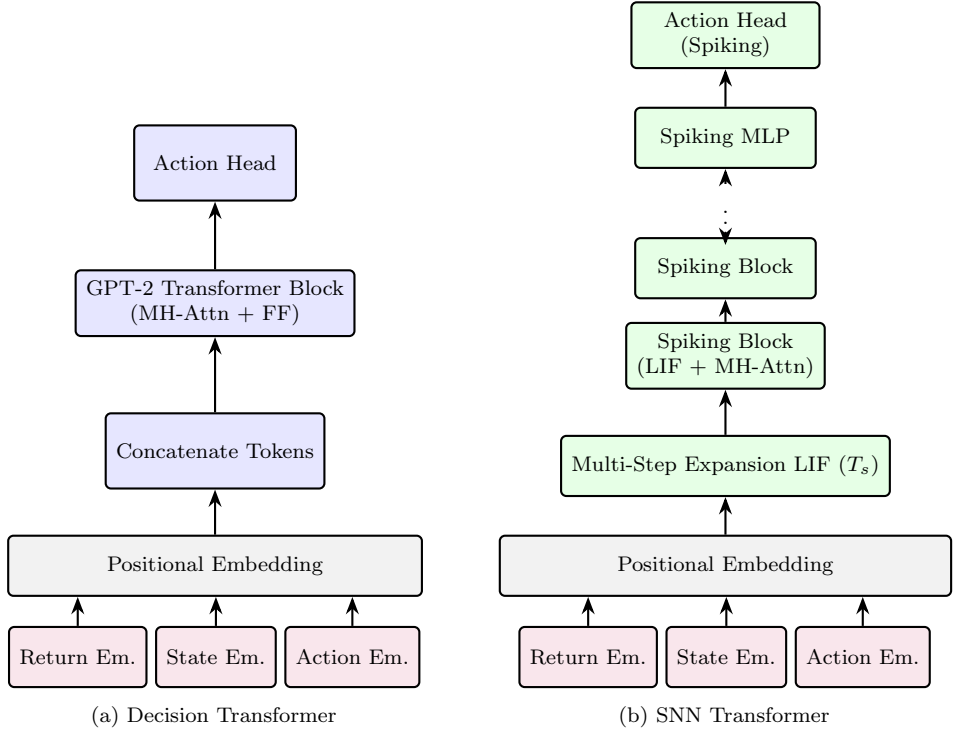
\begin{figure}
\centering
\vspace{-7em}         
%
%----------------- (A) DECISION TRANSFORMER -----------------%
\begin{subfigure}[t]{0.49\textwidth}
\centering
\begin{tikzpicture}[
    font=\scriptsize,
    node distance=0.8cm,
    every node/.style={rounded corners=2pt, align=center},
    layer/.style={draw, thick, rectangle, fill=blue!10, minimum width=2.15cm, minimum height=1.0cm},
    embed/.style={draw, thick, rectangle, fill=purple!10, minimum width=1cm, minimum height=0.8cm},
    attention/.style={draw, thick, rectangle, fill=blue!10, minimum width=2.4cm, minimum height=0.8cm},
    arr/.style={-{Stealth}, thick},
    ln/.style={draw, thick, rectangle, fill=gray!10, minimum width=5.5cm, minimum height=0.8cm}
]

% =========== Embeddings row =========== %
\node[embed] (returns) {Return Em.};
\node[embed, right=0.1cm of returns] (states) {State Em.};
\node[embed, right=0.1cm of states] (actions) {Action Em.};

% Positional Embeddings
\node[ln, above=0.8cm of $(returns)!0.5!(actions)$] (posE) {Positional Embedding};

\draw[arr] (returns.north) -- (posE.south -| returns);
\draw[arr] (states.north) -- (posE.south -| states);
\draw[arr] (actions.north) -- (posE.south -| actions);

% Token Stack
\node[layer, above=0.6cm of posE] (stack) {Concatenate Tokens};
\draw[arr] (posE) -- (stack);

%  Transformer Block
\node[attention, above=1.0cm of stack] (block1) {GPT-2 Transformer Block\\(MH-Attn + FF)};
\draw[arr] (stack) -- (block1);

% Output Head
\node[layer, above=0.9cm of block1] (head) {Action Head};
\draw[arr] (block1) -- (head);
\end{tikzpicture}
\caption{Decision Transformer}
\label{fig:dt_subfigure}
\end{subfigure}
%\hfill
%---------------- (B) SNN TRANSFORMER ----------------%
\begin{subfigure}[t]{0.48\textwidth}
\centering
\begin{tikzpicture}[
    font=\scriptsize,
    node distance=0.8cm,
    every node/.style={rounded corners=2pt, align=center},
    box/.style={draw, thick, rectangle, fill=green!10, minimum width=2.5cm, minimum height=.8cm},
    embed/.style={draw, thick, rectangle, fill=purple!10, minimum width=1cm, minimum height=0.8cm},
    attention/.style={draw, thick, rectangle, fill=green!10, minimum width=2.4cm, minimum height=0.8cm},
    arr/.style={-{Stealth}, thick},
    ln/.style={draw, thick, rectangle, fill=gray!10, minimum width=6cm, minimum height=0.8cm}
]

% ========== Embeddings row ========== %
\node[embed] (returnsSNN) {Return Em.};
\node[embed, right=0.1cm of returnsSNN] (statesSNN) {State Em.};
\node[embed, right=0.1cm of statesSNN] (actionsSNN) {Action Em.};

% Positional Embedding
\node[ln, above=0.8cm of $(returnsSNN)!0.5!(actionsSNN)$] (posESNN) {Positional Embedding};

\draw[arr] (returnsSNN.north) -- (posESNN.south -| returnsSNN);
\draw[arr] (statesSNN.north)  -- (posESNN.south -| statesSNN);
\draw[arr] (actionsSNN.north) -- (posESNN.south -| actionsSNN);

% Multi-step expansion
\node[box, above=0.5cm of posESNN] (expansion) {Multi-Step Expansion LIF ($T_s$)};
\draw[arr] (posESNN) -- (expansion);

% Spiking Transformer Blocks
\node[attention, above=0.6cm of expansion] (sblock1) {Spiking Block\\(LIF + MH-Attn)};
\node[attention, above=0.3cm of sblock1] (sblock2) {Spiking Block};
\node (sdots) at ($(sblock2.north)+(0,0.3cm)$) {\(\vdots\)};
\node[attention, above=0.2cm of sdots] (sblockN) {Spiking MLP};

\draw[arr] (expansion) -- (sblock1);
\draw[arr] (sblock1) -- (sblock2);
\draw[arr] (sblock2) -- (sdots);
\draw[arr] (sdots) -- (sblockN);

% Output Head
\node[box, above=0.5cm of sblockN] (headSNN) {Action Head\\(Spiking)};
\draw[arr] (sblockN) -- (headSNN);

\end{tikzpicture}
\caption{SNN Transformer}
\label{fig:snn_subfigure}
\end{subfigure}

%\vspace{-0.5em}
\caption{The improvement from Decision Transformer (\subref{fig:dt_subfigure}) to SNN Transformer (\subref{fig:snn_subfigure}).
The SNN Transformer replaces dense activations with multi-step LIF neurons, yielding a spike-driven attention mechanism that integrates signals across $T_s$ micro-timesteps.}
\label{fig:dt_vs_snn_twocol}
\vspace{1em}
\end{figure}

% \subsection{Model Construcction}
% \label{sec:model_architecture}
The model combines (i) the representational capacity of Transformers for long-horizon sequence modeling\citep{NIPS2017_3f5ee243} and (ii) the biologically inspired, energy-efficient dynamics of spiking neural networks\citep{neftci2019surrogategradientlearningspiking,Roy2019TowardsSM}. Specifically, a \textbf{SNN Transformer} that processes state, action, return-to-go, and timestep embeddings in a multi-step  fashion is introduced. This section describes each component of the model (Figure~\ref{fig:dt_vs_snn_twocol}(b)) in detail and provides the associated mathematical formulation.
% Insert the full-page image spanning both columns
%\begin{figure*}[!htbp]
  %\centering
  %\includegraphics[width=\textwidth,height=1\textheight,keepaspectratio]{arc.png}
  %\caption{Model Architecture Overview}
  %\label{fig:arc_full}
%\end{figure*}

\subsection{ Input Representations}

Following the trditional DRL design \citep{712192}, let the agent be described by a tuple $(s_t, a_{t-1}, G_t)$, where \(\mathbf{s}_t \in \mathcal{S}\) is the agent’s state, \(a_{t-1} \in \mathcal{A}\) is the agent's previous action, \(G_t \in \mathbb{R}\) is the return-to-go.

These components are then mapped into a common embedding dimension \(d\). Specifically,
\[
\begin{aligned}
\mathbf{e}^{(s)}_t &= W_s \, \mathbf{s}_t, \quad
\mathbf{e}^{(a)}_t = W_a \, \mathrm{one\_hot}(a_{t-1}),\\
\mathbf{e}^{(r)}_t &= W_r \, G_t, \quad
\mathbf{e}^{(t)}_t = E_t(t),
\end{aligned}
\]

where \(W_s \in \mathbb{R}^{d\times d_s}\), \(W_a \in \mathbb{R}^{d \times |\mathcal{A}|}\), and \(W_r \in \mathbb{R}^{d \times 1}\) are learnable projection matrices, while \(E_t(\cdot)\) is an embedding for the integer timestep. The function \(\operatorname{one\_hot}(\cdot)\) converts the discrete action label into a length-\(|\mathcal{A}|\) vector that is all zeros except for a single \(1\) at the index corresponding to \(a_{t-1}\). A learnable positional embedding \(\mathbf{p}\in \mathbb{R}^{T_{\max}\times d}\) is also added to each token.

\subsection{Multi-Step LIF Dynamics}
In the state‑of‑the‑art Decision Transformer algorithm \citep{chen2021decisiontransformerreinforcementlearning}, each linear projection or multi‑head‑attention output is immediately processed by layer normalisation, a \textsc{gelu} non‑linearity, and a residual addition.  All three operations act \emph{instantaneously} on dense real‑valued activations; the hidden state is therefore recomputed from scratch at every time step with no intrinsic temporal memory.  While this design achieves strong performance on GPUs, it entails millions of floating‑point multiply–accumulate operations per token and offers no sparsity that could be exploited by neuromorphic hardware.  Moreover, because information is refreshed rather than integrated across micro‑steps, the model must learn to re‑encode long‑range dependencies at every layer—an energetically inefficient strategy for long‑horizon tasks.  To overcome these limitations, we replace the post‑projection activation with a multi‑step spiking neuron.  

Unlike standard Transformers, each linear or attention operation in the architecture is followed by a \textbf{multi-step spiking neuron}---a Leaky Integrate-and-Fire (LIF) unit that integrates synaptic current over discrete timesteps \(\tau \in \{1,\dots,T_{\mathrm{max}}\}\) and emits spikes when the membrane potential surpasses a threshold. A simplified LIF neuron \citep{neftci2019surrogategradientlearningspiking,fang2021incorporatinglearnablemembranetime} satisfies:
\begin{equation}
U_{i,t+1}^\ell = \alpha \,U_{i,t}^\ell + I_{i,t}^\ell - R_{i,t}^\ell
\end{equation}
where \(U_{i,t}^\ell\) is the membrane potential of neuron \(i\) at time \(\tau=t\) in layer \(\ell\); \(\alpha \in (0,1)\) is a leak factor, \(I_{i,t}^\ell\) is the synaptic current (output from a linear or attention layer), and \(R_{i,t}^\ell\) is a reset term. The neuron emits a spike \(S_{i,t}^\ell\) if \(U_{i,t+1}^\ell>\theta\), typically resetting the membrane potential:
\[
S_{i,t}^\ell = \mathbb{H}(U_{i,t+1}^\ell - \theta), 
\quad
R_{i,t}^\ell = U_{i,t+1}^\ell \cdot S_{i,t}^\ell,
\]
where \(\mathbb{H}(\cdot)\) is the Heaviside step function and \(\theta\) is the firing threshold. In practice, a surrogate gradient approach for backpropagation through the non-differentiable spike function is adopted\citep{neftci2019surrogategradientlearningspiking,bellec2020solution}.

\subsection{Self-Attention with LIF}

Each \textbf{SNN Transformer Block} begins with a spiking multi-head self-attention mechanism\citep{li2022spikeformernovelarchitecturetraining,zhou2022spikformerspikingneuralnetwork}. For a given input sequence \(\{\mathbf{x}_1,\dots,\mathbf{x}_S\}\in \mathbb{R}^{S\times d}\), linear projections are used to obtain queries \(\mathbf{Q}\), keys \(\mathbf{K}\), and values \(\mathbf{V}\) :
\begin{equation}
\mathbf{Q} = \mathbf{X} \, W_Q,\quad 
\mathbf{K} = \mathbf{X} \, W_K,\quad 
\mathbf{V} = \mathbf{X} \, W_V,
\end{equation}
where \(\mathbf{X}\) is the matrix of embeddings concatenated along the sequence dimension. Each projection is then passed through a multi-step LIF layer:
\begin{equation}
\mathbf{Q}_{\mathrm{s}} = \mathrm{LIF}(\mathbf{Q}),\quad
\mathbf{K}_{\mathrm{s}} = \mathrm{LIF}(\mathbf{K}),\quad
\mathbf{V}_{\mathrm{s}} = \mathrm{LIF}(\mathbf{V})
\end{equation}
\(\mathbf{Q}_{\mathrm{s}}, \mathbf{K}_{\mathrm{s}}, \mathbf{V}_{\mathrm{s}}\) are split into \(h\) heads of dimension \(d/h\), and the scaled dot-product attention is computed:
\begin{equation}
\mathrm{Attn}(\mathbf{Q}, \mathbf{K}, \mathbf{V})
= 
\mathrm{softmax}\!\Bigl(\tfrac{\mathbf{Q}_{\mathrm{spike}}\, \mathbf{K}_{\mathrm{spike}}^{\top}}{\sqrt{d/h}}\Bigr)\,
\mathbf{V}_{\mathrm{spike}}
\end{equation}
The resulting attention output is passed through a final LIF neuron and aggregated over \(T_{\mathrm{max}}\) time steps (e.g., by taking the mean activation).

\subsection{Spiking MLP Sub-layer}

Following the attention sub-layer and a residual connection, layer normalization is applied and then a \textbf{spiking MLP} sub-layer consisting of:
\begin{enumerate}
    \item A linear projection from dimension \(d\) to a larger hidden dimension \(d_{\mathrm{hidden}}\).
    \item A multi-step LIF activation.
    \item A second linear projection back to dimension \(d\).
    \item Another multi-step LIF activation.
\end{enumerate}

Formally, for an intermediate representation \(\mathbf{z}\in\mathbb{R}^{S\times d}\):
\[
\mathbf{h} = \mathrm{LIF}(\mathbf{z}\,W_1 + \mathbf{b}_1), 
\quad 
\mathbf{y} = \mathrm{LIF}(\mathbf{h}\,W_2 + \mathbf{b}_2),
\]
where \(W_1,W_2\) and \(\mathbf{b}_1,\mathbf{b}_2\) are trainable parameters. An additional skip connection adds \(\mathbf{z}\) to \(\mathbf{y}\), implementing the usual Transformer block structure with spiking non-linearities.

\subsection{Final Output Head}

After passing through \(L\) stacked \textbf{SNN Transformer Blocks}, the output representation \(\mathbf{X}^{(L)} \in \mathbb{R}^{S\times d}\) is fed into an action prediction head. A simple linear readout projects each token’s embedding to logits over the action space \(\mathcal{A}\):
\begin{equation}
\hat{\mathbf{a}}_t = \mathbf{w}_{\text{out}}^{\top} \,\mathbf{x}^{(L)}_t + \mathbf{b}_{\text{out}}
\end{equation}
where \(\hat{\mathbf{a}}_t \in \mathbb{R}^{|\mathcal{A}|}\) represents unnormalized probabilities for the four discrete directions \(\{\text{left},\text{right},\text{up},\text{down}\}\). We apply a cross-entropy loss over valid (non-padded) tokens to train the parameters \(\theta\) of the entire network end-to-end.

Overall, this \emph{SNN Transformer} couples the power of attention-based sequence modeling with the biologically inspired design of spiking neural networks, thereby offering an energy-efficient and scalable framework for reinforcement learning in complex, long-horizon tasks.

\subsection{Improvements from Decision Transformer}
\label{sec:theory_advantages}

We consider a general sequence modeling problem where the input is a sequence \(X = \{x_1, x_2, \dots, x_T\}\) and the objective is to predict a sequence of actions \(A = \{a_1, a_2, \dots, a_T\}\). In a standard Decision Transformer (DT) Figure~\ref{fig:dt_vs_snn_twocol}(a), each token is processed by a series of dense, feedforward layers and self-attention blocks. Mathematically, the hidden representation at time \(t\) is given by
\[
h_t = f(x_t),
\]
and the self-attention mechanism computes
\[
\text{Attention}(Q, K, V) = \operatorname{softmax}\left(\frac{QK^\top}{\sqrt{d}}\right)V,
\]
where \(Q = W_Q h_t\), \(K = W_K h_t\), \(V = W_V h_t\), and \(d\) is the embedding dimension.

In contrast, the SNN Transformer replaces dense activations with spiking neuron dynamics. At each layer, the input is processed by a multi-step Leaky Integrate-and-Fire (LIF) neuron. Let \(U_t\) denote the membrane potential at time \(t\) and consider the following update:
\[
U_t = \alpha U_{t-1} + f(x_t) - R_t,
\]
where \(\alpha \in (0,1)\) is a leak factor and \(R_t\) is the reset term when \(U_t\) exceeds a threshold \(\theta\). The spiking activation is then defined as
\[
S_t = \mathbb{H}(U_t - \theta),
\]
with \(\mathbb{H}(\cdot)\) being the Heaviside step function. Over a fixed number of simulation steps \(T_s\), the effective output is an average of these spikes:
\[
h_t^{\text{SNN}} = \frac{1}{T_s}\sum_{\tau=1}^{T_s} S_t^{(\tau)}.
\]
This multi-step integration acts as a temporal smoothing filter that can be mathematically interpreted as a convolution with an exponentially decaying kernel (due to the leak factor \(\alpha\)), thereby retaining salient features over extended time horizons.

For the self-attention mechanism in the SNN Transformer, the dense queries, keys, and values are replaced by their spiking counterparts:
\[
\tilde{Q} = \frac{1}{T_s}\sum_{\tau=1}^{T_s} \mathbb{H}(Q^{(\tau)} - \theta_Q), \quad
\tilde{K} = \frac{1}{T_s}\sum_{\tau=1}^{T_s} \mathbb{H}(K^{(\tau)} - \theta_K), \quad
\tilde{V} = \frac{1}{T_s}\sum_{\tau=1}^{T_s} \mathbb{H}(V^{(\tau)} - \theta_V).
\]
The spiking self-attention is then computed as
\[
\text{Attention}_{\text{SNN}}(\tilde{Q}, \tilde{K}, \tilde{V}) = \operatorname{softmax}\!\left(\frac{\tilde{Q}\tilde{K}^\top}{\sqrt{d}}\right)\tilde{V}.
\]
Due to the thresholding operation \(\mathbb{H}(\cdot)\), only the most salient activations contribute to \(\tilde{Q}\), \(\tilde{K}\), and \(\tilde{V}\). The integration over \(T_s\) simulation steps further enhances the representation by averaging out transient noise and emphasizing persistent signals. 

Thus, while the Decision Transformer computes representations in a memoryless, dense manner:
\[
h_t = f(x_t),
\]
the SNN Transformer computes
\[
h_t^{\text{SNN}} = \frac{1}{T_s}\sum_{\tau=1}^{T_s} \mathbb{H}\Bigl(\alpha U_{t-1}^{(\tau)} + f(x_t) - \theta\Bigr),
\]
which can be seen as a form of adaptive, event-driven integration. This integration confers two principal advantages: (1) enhanced robustness to noise by filtering out minor fluctuations through thresholding and averaging, and (2) improved capacity to capture extended temporal dependencies, since the membrane potential retains information from prior timesteps. These properties are not only beneficial in grid-based maze navigation but also generalize to other sequential decision-making tasks where long-range dependencies and noise robustness are critical. Consequently, the SNN Transformer exhibits superior performance compared to the Decision Transformer, as evidenced by empirical results in Section~\ref{subsec:compare_DT} where accuracy increases from approximately 80\% in DT to over 99\% in the SNN Transformer.

\subsection{Spiking Transformer-Based Learning}
\label{subsec:spiking_transformer_rl}

 Similar to the decision transformer algorithm \citep{chen2021decisiontransformerreinforcementlearning}, expert dmonstrations are required to train the Spike-transformer network. Let the list of expert demonstration be encoded as a sequence of tokens \(\{(\mathbf{s}_t, a_{t-1}, G_t, t)\}_{t=1}^T\), where \(a_0\) is a dummy token for the initial step. The spiking attention and MLP sub-layers process these embedded tokens over multi-step LIF neurons, capturing both:

\begin{itemize}
    \item \textbf{Long-Range Dependencies.} Self-attention attends to relevant positions throughout the trajectory, crucial for pathfinding tasks.
    \item \textbf{Neuro-Inspired Efficiency.} Multi-step spiking neurons can exploit sparse firing, offering potentially lower computational cost on neuromorphic hardware.
\end{itemize}

The network outputs logits \(\hat{\mathbf{a}}_t\) over the actions space $\mathcal{A}$ at each timestep \(t\). The spike transformer network is then optimized by the following loss function:
\begin{equation}
\mathcal{L}_{\text{CE}} = - \sum_{t=1}^T 
    \left( \log \, p_\theta\left(a_t \,\middle|\, \mathbf{s}_{1:t}, a_{1:t-1}, G_{1:t}, t_{1:t}\right) \right) \cdot \mathbb{I}_{\{ t \text{ not padded} \}}
\label{eq:cross_entropy_loss}
\end{equation}
where \(\mathbb{I}\{\cdot\}\) is an indicator function ignoring padded positions in trajectories that are shorter than a maximum length \(S\).

%The method unifies \emph{spiking neural networks} (SNNs) with the representational power of \emph{Transformer} architectures to handle extended sequence modeling in offline RL:
\section{Experiments}
\label{sec:experiments}
\label{subsec:data_snn_training}

\subsection{Experiment Setup}
\label{subsec:data_gen}
To verify the performance of the developed STRL algorithm, a extensive empirical experiments were conducted on a popular test bench, i.e., the maze navigation problem from D4RL \citep{fu2021d4rldatasetsdeepdatadriven}. This section outlines the experimental design, including data splits and training procedure, followed by quantitative benchmarks and qualitative analyses of predicted paths.

\noindent{In the maze experiment,} STRL is evaluated on two complementary environments.  
First, in \emph{procedurally generated \(21{\times}21\) grid mazes}, a depth-first backtracker carves a single-solution labyrinth where walls (1) and corridors (0) alternate in a checkerboard pattern.  
The agent starts at \((0,1)\) just inside the western wall and must reach \((W-1,H-2)\) adjacent to the eastern wall.  
Its state is the integer coordinate \((x_t,y_t)\) and the action set is \{\textit{left},\textit{right},\textit{up},\textit{down}\}; attempting to step into a wall leaves the position unchanged.  
Each move incurs a \(-0.1\) penalty, reaching the goal yields \(+1.0\), and the episode is truncated after \(T_{\max}=100\) steps.  
A* search provides the unique shortest path, and the resulting \((\text{state},\text{action},\text{reward})\) triplets form the expert trajectories used for training.  

\smallskip
Second, the continuous \emph{D4RL \texttt{maze2d-umaze-v1}} environment is a U-shaped corridor with positions in \([-1,1]^2\).  
Logged two-dimensional velocity commands are discretised to the same four cardinal actions by taking the dominant sign of each component, while the observation passed to STRL remains the raw \((x_t,y_t)\) position.  
Rewards supplied by the dataset combine shaped forward progress with a terminal bonus; we post-process each trajectory to compute per-step returns-to-go.  
Training is entirely offline—the agent never interacts with the environment but learns solely from these fixed demonstrations.  

\smallskip
Together, these mazes stress distinct aspects of long-horizon control: combinatorial reasoning in the discrete grids and precise continuous navigation in \texttt{maze2d}.  
SNN Transformer is trained and evaluated on each environment independently.

For each maze, we run an A*\cite{4082128} solver to extract the shortest  path from start to goal. The resulting trajectory provides:
\begin{itemize}
    \item The sequence of states \(\{\mathbf{s}_1, \mathbf{s}_2, \dots, \mathbf{s}_T\}\).
    \item The corresponding actions \(\{a_1, a_2, \dots, a_T\}\).
    \item Step-based rewards \(\{r_1, r_2, \dots, r_T\}\), combining negative step penalties and a terminal bonus.
\end{itemize}
The process to compute returns-to-go \(G_t\) is by summing from time \(t\) until the episode ends, effectively labeling each state with a future reward estimate.

The dataset comprises 50,000 randomly generated $21 \times 21$ mazes, each containing a single optimal or near-optimal trajectory from a start to a goal cell via A* search. Each solution path is labeled with (state, action, reward) triplets, where rewards consist of small negative step costs ($-0.1$) plus a terminal reward ($+1$). This dataset is split into 70\% training, 15\% validation, and 15\% test sets, ensuring no overlap in maze layouts across splits. States are normalized per dimension to mitigate scale disparities.

We also normalize each state by subtracting the mean and dividing by the standard deviation, computed over the training portion of the procedural dataset,
\[
\tilde{\mathbf{s}}_t \;=\;
\frac{\mathbf{s}_t \;-\; \boldsymbol{\mu}}
     {\boldsymbol{\sigma}}
\quad\in\mathbb{R}^{d_s},
\tag{18}
\]
\noindent where
\[
\mathbf{s}_t = [\,x_t,\;y_t\,]^\top,
\quad
\boldsymbol{\mu} = \frac{1}{N}\sum_{n=1}^N\mathbf{s}^{(n)},
\]
\[
\boldsymbol{\sigma} = \sqrt{\frac{1}{N}\sum_{n=1}^N\bigl(\mathbf{s}^{(n)}-\boldsymbol{\mu}\bigr)^2}\;+\;\varepsilon,.
\]
and a small \(\varepsilon\) is added to \(\boldsymbol{\sigma}\) to avoid division by zero. The same normalization is applied to the D4RL mazes for consistency. We then split each dataset into training, validation, and test sets, ensuring coverage of diverse maze configurations.

\subsection{STRL Design}
The STRL structure consists of six SNN Transformer blocks stacked together. And each interleaving spiking self-attention and a spiking MLP. The embedding dimension is 256, and all spiking neurons use multi-step LIF nodes with $T = 4$ time steps. Positional embeddings and embeddings for state, action, return-to-go, and timesteps are added to form the input tokens for the self-attention mechanism. 

When training the SNN Transformer, the AdamW optimizer \citep{loshchilov2019decoupledweightdecayregularization} with a cosine-annealing schedule \citep{loshchilov2017sgdrstochasticgradientdescent} was utilized. The gradients are clipped at a norm of 1.0 to prevent exploding updates. Each training batch randomly samples from  both procedural and D4RL trajectories,ensuring coverage of diverse maze structures.

The proposed STRL network is trained for 10 epochs. Each training epoch consumes mini-batches of size 32, randomly sampling from the set of truncated or padded trajectories (up to 100 steps). The network is optimized with \textit{AdamW} (\(\mathrm{lr} = 10^{-3}\), weight decay $=10^{-4}$) using a cosine annealing schedule. To stabilize updates, gradients are clipped at a norm of 1.0. We measure performance via cross-entropy loss on correctly predicting each action within the trajectory, masking any padded positions.

\subsection{Results Analysis}
We track accuracy on the validation set to tune hyperparameters such as learning rate, embedding dimension \(d\), and the number of Transformer layers \(L\). We evaluate training, validation, and test performance every epoch. At each epoch, we measure both \emph{per-step action accuracy} and \emph{path fidelity}, comparing predicted paths to the A* solutions in procedural mazes and the ground-truth references in D4RL. 

\begin{figure}
\vspace{-3em}
\centering
\includegraphics[width=.48\linewidth]{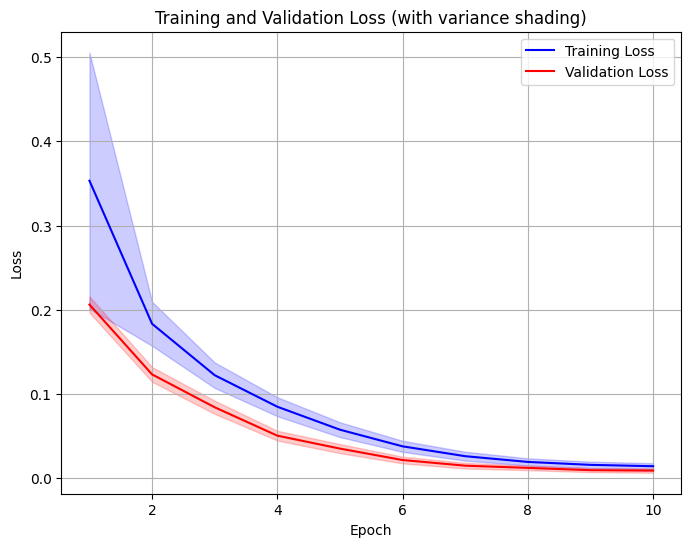}
\quad
\includegraphics[width=.48\linewidth]{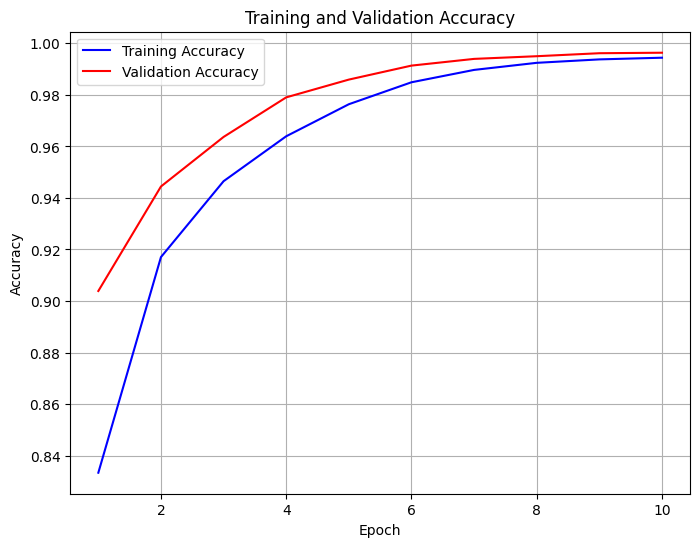}
\caption{(\textbf{Top}) Training and validation loss, showing mean and standard deviation per epoch. (\textbf{Bottom}) Training and validation accuracy. Both curves illustrate stable convergence and close alignment between training and validation sets.}
\label{fig:loss_acc_plots}
\end{figure}

\paragraph{Loss and Accuracy}
Figure~\ref{fig:loss_acc_plots} (top) depicts the training and validation loss trajectories with variance shading over mini-batch losses in each epoch. After a sharp drop in the initial epochs, the model refines steadily, reaching near-zero loss on both sets. The accuracy curves in Figure~\ref{fig:loss_acc_plots} (bottom) similarly exhibit rapid improvement from 83\% to beyond 99\%, with minimal gap between training and validation performance. These observations indicate that spiking-based attention effectively learns from offline demonstration data and avoids substantial overfitting.

\paragraph{Convergence and Learning Curves}

Table~\ref{tab:epoch_metrics} summarizes epoch-wise training and validation outcomes, showing rapid improvement over the first several epochs. By epoch 3, the model surpasses 94\% accuracy on training data and 96\% on the validation set. Convergence continues steadily, reaching $>99\%$ on both sets by epoch 7. Ultimately, the validation accuracy peaks at 99.64\% by epoch 10.

\begin{table}[h]
%\vspace{-2em}
\centering
\begin{tabular}{lccccc}
\toprule
\textbf{Epoch} & \textbf{Train Loss} & \textbf{Train Acc} & \textbf{Val Loss} & \textbf{Val Acc} \\
\midrule
1 & 0.3533 & 83.34\% & 0.2063 & 90.39\% \\
2 & 0.1834 & 91.71\% & 0.1234 & 94.45\% \\
3 & 0.1224 & 94.65\% & 0.0843 & 96.37\% \\
4 & 0.0850 & 96.39\% & 0.0507 & 97.90\% \\
5 & 0.0578 & 97.64\% & 0.0353 & 98.59\% \\
6 & 0.0380 & 98.49\% & 0.0218 & 99.14\% \\
7 & 0.0263 & 98.97\% & 0.0151 & 99.40\% \\
8 & 0.0195 & 99.24\% & 0.0124 & 99.50\% \\
9 & 0.0160 & 99.38\% & 0.0097 & 99.62\% \\
10 & 0.0144 & 99.44\% & 0.0092 & 99.64\% \\
\bottomrule
\end{tabular}
\caption{Epoch-level metrics during training. The model converges rapidly to above 99\% accuracy on both training and validation sets.}
\label{tab:epoch_metrics}
\end{table}

\paragraph{Confusion Analysis}

Selection of the best checkpoint is based on validation accuracy and evaluated on the held-out test set of mazes. The model achieves a \textbf{test loss of 0.0090} and \textbf{test accuracy of 99.64\%}. This test performance confirms robust generalization to previously unseen maze layouts.

\begin{figure}[h]
%\vspace{-4em}
\centering
\includegraphics[width=.7\linewidth]{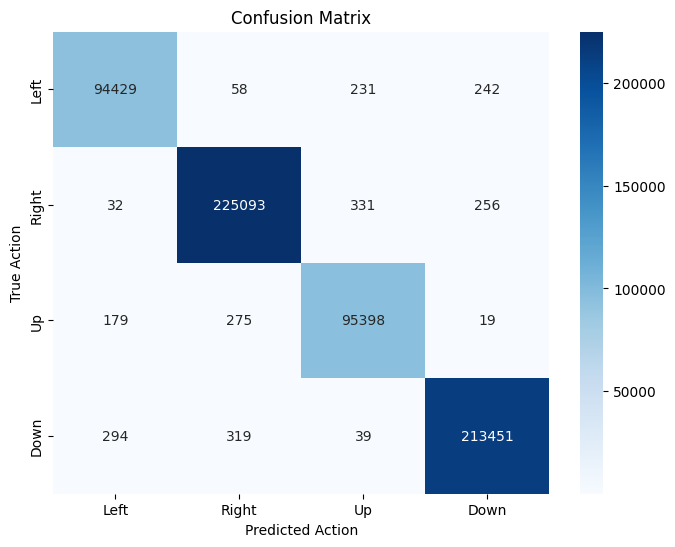}
\caption{Test confusion matrix over 4 actions. The model maintains high recall and precision for each class, evidencing minimal misclassifications.}
\label{fig:confmat}
\end{figure}

Figure~\ref{fig:confmat} shows the confusion matrix in the test set. The diagonal dominance highlights near-perfect classification of all four moves (left, right, up, down). Off-diagonal entries remain exceptionally small, indicating that even subtle differences between adjacent actions (e.g., left vs. up) are readily distinguished by the spiking attention module.

\paragraph{Comparison with Decision Transformer}
\label{subsec:compare_DT}

A series of comparison experiments are also conducted to compare STRL's performance with Decision Transformer\citep{chen2021decisiontransformerreinforcementlearning}.

\begin{figure}
\vspace{-3em}
\centering
\includegraphics[width=.48\linewidth]{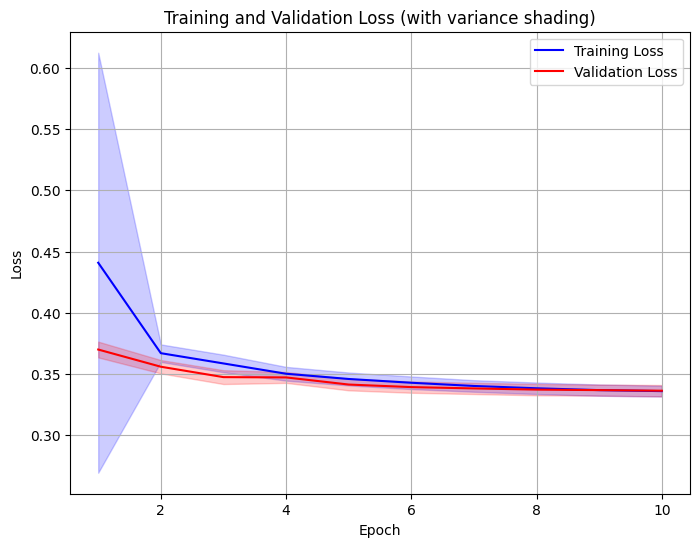}
\quad
\includegraphics[width=.48\linewidth]{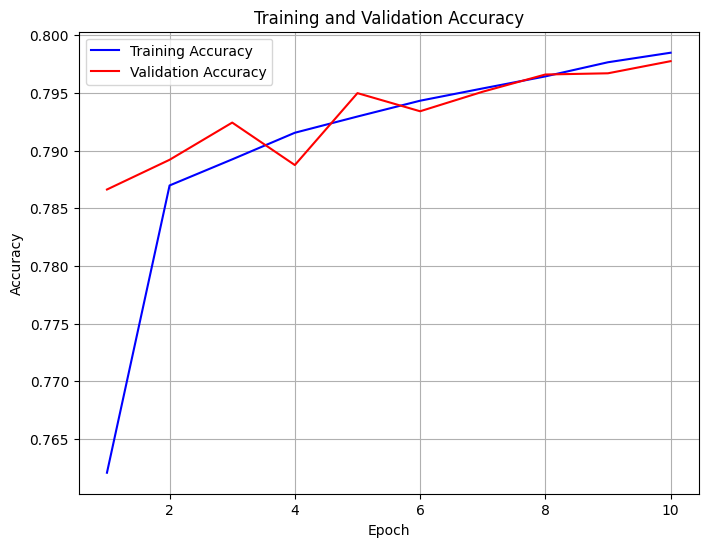}
\caption{(\textbf{Left}) Decision Transformer training and validation loss, showing mean and standard deviation per epoch. (\textbf{Right}) Training and validation accuracy for Decision Transformer. The model exhibits stable convergence but achieves lower final accuracy compared to the SNN Transformer.}
\label{fig:dt_loss_acc_plots}
\end{figure}

Figure~\ref{fig:dt_loss_acc_plots} illustrates the training and validation accuracy/loss curves for the Decision Transformer across 10 epochs, while Table~\ref{tab:compare_dt} summarizes the final test metrics relative to the proposed SNN Transformer.

As shown in Table~\ref{tab:compare_dt}:

\begin{itemize}
    \item \textbf{Decision Transformer Results:}  
    The Decision Transformer achieved a training accuracy of $\sim79.9\%$ after 10 epochs and a test accuracy of $79.82\%$, with a test loss of $0.3357$. These trends are also reflected in Figure~\ref{fig:dt_loss_acc_plots}(Top), where both training and validation curves converge to around $0.80$ accuracy and $0.34$ loss.

    \item \textbf{SNN Transformer Results:}  
    The proposed method converged to a test accuracy of $99.64\%$, and a test loss of $0.0090$, indicating more precise action predictions over the entire maze navigation dataset.
\end{itemize}

\begin{table}[h]
\centering
\begin{tabular}{lcc}
\toprule
\textbf{Model} & \textbf{Test Loss} & \textbf{Test Accuracy} \\
\midrule
Decision Transformer (DT) & 0.3357 & 79.82\% \\
SNN Transformer (Ours) & 0.0090 & 99.64\% \\
\bottomrule
\end{tabular}
\caption{Comparison of Decision Transformer vs. SNN Transformer on the same offline maze dataset of $50{,}000$ samples.}
\label{tab:compare_dt}
\end{table}

\subsection{Additional Results on D4RL maze dataset}
\label{subsec:d4rl_results}

While our primary experiments focus on the $21\times 21$ A* dataset, we further validate our approach on the D4RL \texttt{maze2d-umaze-v1} dataset~\citep{fu2021d4rldatasetsdeepdatadriven}, which provides continuous $(x,y)$ states and transition data. We approximate the four discrete actions (left, right, up, down) by computing the dominant direction between consecutive positions. This yields a set of trajectories that we split into $80\%$ training and $20\%$ validation. We maintain the same hyperparameters (embedding dimension $=256$, multi-step LIF nodes with $T=4$) and optimizer settings used in our A* experiments, but train for 20 epochs due to the dataset’s larger variability in continuous-state transitions.
\begin{figure}[h]
\centering
\includegraphics[width=.85\linewidth]{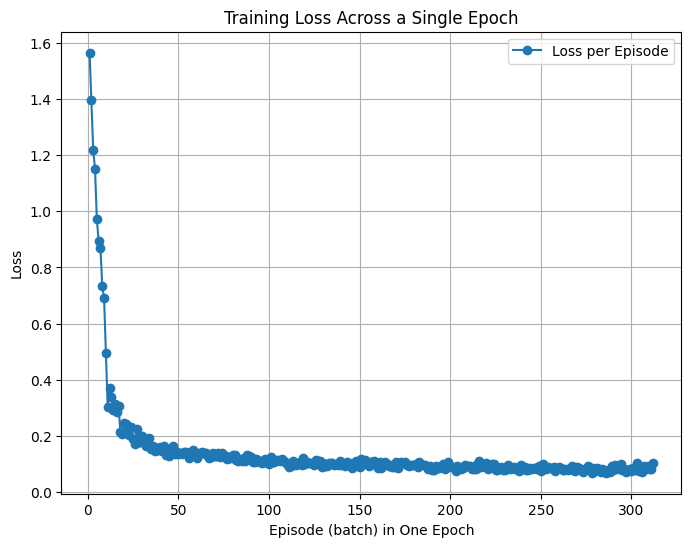}
\caption{ Training loss per episode (mini-batch) for a single epoch on \texttt{maze2d-umaze-v1}. The SNN Transformer converges rapidly within the first 50 batches.}
\label{fig:F1}
\end{figure}

\begin{figure}[!ht]
  \centering
  % left: loss plot
  \begin{subfigure}[b]{0.48\textwidth}
    \includegraphics[width=\linewidth]{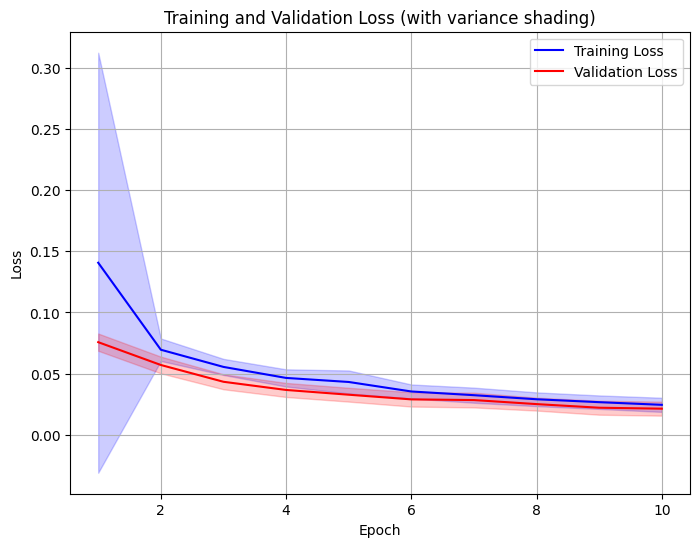}
  \end{subfigure}\hfill
  % right: accuracy plot
  \begin{subfigure}[b]{0.48\textwidth}
    \includegraphics[width=\linewidth]{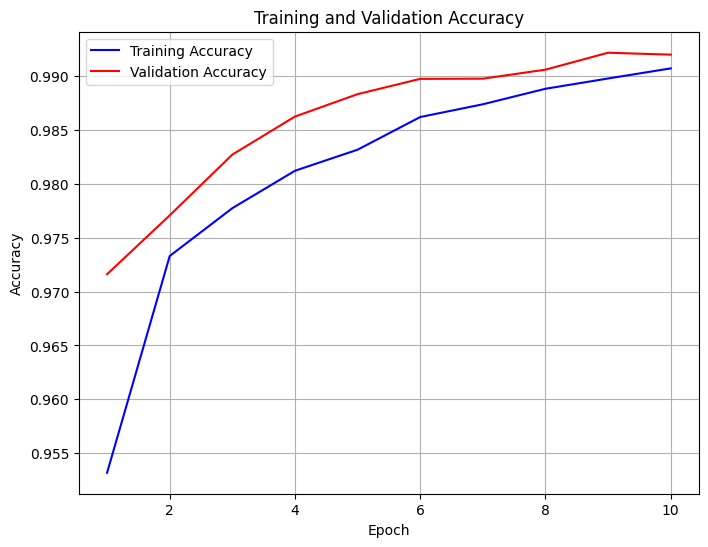}
  \end{subfigure}
  \caption{%
    \textbf{(Left)} Training (blue) and validation (red) loss with variance shading. The loss drops below 0.01 by epoch 15, demonstrating stable convergence.%
    \quad%
    \textbf{(Right)} Training (blue) and validation (red) accuracy on \texttt{maze2d-umaze-v1} across 20 epochs. The model surpasses 99\% accuracy after only a few epochs.%
  }
  \label{fig:loss_and_acc}
\end{figure}

\paragraph{Learning Curves.}
Figure~\ref{fig:F1} illustrates the \emph{per-episode} training loss within the \emph{first epoch}, measured at each mini-batch. The model’s cross-entropy drops sharply from around $1.6$ down below $0.2$ in fewer than 50 batches, demonstrating rapid adaptation to the D4RL trajectories. Figure~\ref{fig:loss_and_acc} then depict the epoch-level convergence across all 10 epochs. Specifically:
\begin{itemize}
    \item \textbf{Figure~\ref{fig:loss_and_acc}} \textbf{(Right)} (Training and Validation Accuracy) shows a swift ascent from around $95\%$ at epoch~1 to nearly $99.5\%$ by epoch~10, culminating in $99.55\%$ (training) and $99.59\%$ (validation) accuracy at epoch~10.
    \item \textbf{Figure~\ref{fig:loss_and_acc}}\textbf{(Left)} (Training and Validation Loss) highlights the steady decline in cross-entropy, with validation loss consistently tracking close to training loss. By epoch~5, both curves dip below $0.05$, and after epoch~10, they approach near-zero values with minimal variance, indicating robust generalization and negligible overfitting.
\end{itemize}

\paragraph{Quantitative Performance.}
Table~\ref{tab:d4rl_epoch_metrics} reports representative metrics at selected epochs. From an initial accuracy near $95\%$, the SNN Transformer rapidly improves to over $99\%$ by epoch~10, ultimately reaching $99.55\%$ on training and $99.59\%$ on validation after 20 epochs. The slight difference between training and validation curves indicates minimal overfitting, corroborating the narrow gap observed in Figure~\ref{fig:loss_and_acc}.

\begin{table}[h!]
\centering
\begin{tabular}{lcccc}
\toprule
\textbf{Epoch} & \textbf{Train Loss} & \textbf{Train Acc} & \textbf{Val Loss} & \textbf{Val Acc}\\
\midrule
1  & 0.1406 & 95.32\% & 0.0758 & 97.16\% \\
5  & 0.0432 & 98.32\% & 0.0328 & 98.83\% \\
10 & 0.0246 & 99.07\% & 0.0214 & 99.20\% \\
\bottomrule
\end{tabular}
\caption{Selected epoch-wise metrics for D4RL \texttt{maze2d-umaze-v1}. The model converges to $\sim99.59\%$ validation accuracy and near-zero cross-entropy.}
\label{tab:d4rl_epoch_metrics}
\end{table}

These results demonstrate that the SNN Transformer adapts seamlessly from the synthetic, discrete A* mazes to the continuous-state trajectories in D4RL, preserving its rapid learning dynamics and achieving near-perfect classification of discrete directions. By epoch 20, the gap between training and validation metrics is consistently below $0.5\%$, underscoring strong generalization capabilities in a more varied real-world-inspired dataset. Together with the findings on the $21\times 21$ mazes, the D4RL analysis further solidifies the SNN Transformer’s suitability for offline RL tasks requiring long-horizon planning, energy efficiency, and accurate spatio-temporal modeling.

Together, these features demonstrate the viability of combining spiking neural components with Transformers to learn complex, long-horizon decision tasks across multiple offline RL datasets.

\subsection{Discussion}

Each experiment with the \textbf{SNN Transformer} was repeated ten times, and \emph{every} run achieved an accuracy exceeding \(99\,\%\).Taken together, these results establish that the spiking-based self-attention and MLP layers can master the spatio-temporal dependencies needed for high-accuracy maze navigation. The smooth loss and accuracy curves, minimal overfitting, and near-perfect confusion matrix reflect the Transformer’s capacity to represent extended trajectories, while the multi-step LIF activations preserve the advantages of event-driven spiking. In practice, such architectures could leverage neuromorphic hardware for more energy-efficient sequential decision-making, with potential applications extending beyond maze-solving to other domains requiring complex path planning or long-horizon RL.

The promising performance of the SNN Transformer motivates further exploration in multiple directions. First, extending the model to continuous control tasks in high-dimensional RL environments could reveal additional benefits of spiking-based sequence modeling. Second, integrating neuromorphic hardware accelerators could enable real-time, energy-efficient decision-making in resource-constrained settings. Third, investigating hybrid spiking-dense architectures may bridge the gap between conventional deep learning and biologically inspired computation, optimizing both performance and efficiency.  

Overall, the findings validate the effectiveness of integrating Transformer-style attention with spiking dynamics, achieving state-of-the-art performance in offline maze navigation and offering a promising direction for future spiking-based RL research.

\section{Conclusions}
\label{sec:conclusion}
This work proposes the \textbf{SNN Transformer} that integrates spiking neural networks (SNNs) with Transformer-based sequence modeling for reinforcement learning (RL) tasks. By leveraging multi-step Leaky Integrate-and-Fire (LIF) neurons within the attention and feedforward layers, the model effectively captures long-range dependencies while benefiting from event-driven computation. The experiments on a comprehensive dataset of 50,000 maze navigation trajectories demonstrate that the SNN Transformer attains a remarkable test accuracy of \textbf{99.64\%}. In comparison, the Decision Transformer achieves a significantly lower accuracy of \textbf{79.82\%} on the same dataset. This notable enhancement highlights the potential of spiking-based architectures to model complex sequential decision-making tasks with exceptional precision. Overall, this work establishes a strong foundation for leveraging SNNs in RL and sequence modeling, highlighting the potential of spiking-based Transformers to advance neuromorphic AI research. The findings suggest that incorporating biologically plausible mechanisms into modern deep learning frameworks can lead to more efficient and scalable solutions for sequential decision-making. In the future, STRL can be ported to next-generation neuromorphic hardware, paving the way for real-time, energy-aware decision-making in autonomous robots and other edge devices.

\bibliography{STRL}

\begin{thebibliography}{10}
\expandafter\ifx\csname url\endcsname\relax
  \def\url#1{\texttt{#1}}\fi
\expandafter\ifx\csname urlprefix\endcsname\relax\def\urlprefix{URL }\fi
\expandafter\ifx\csname href\endcsname\relax
  \def\href#1#2{#2} \def\path#1{#1}\fi

\bibitem{JIANG2021389}
R.~Jiang, Z.~Wang, B.~He, Y.~Zhou, G.~Li, Z.~Zhu, \href{https://www.sciencedirect.com/science/article/pii/S0925231221012078}{A data-efficient goal-directed deep reinforcement learning method for robot visuomotor skill}, Neurocomputing 462 (2021) 389--401.
\newblock \href {https://doi.org/https://doi.org/10.1016/j.neucom.2021.08.023} {\path{doi:https://doi.org/10.1016/j.neucom.2021.08.023}}.
\newline\urlprefix\url{https://www.sciencedirect.com/science/article/pii/S0925231221012078}

\bibitem{LI201992}
F.~Li, Q.~Jiang, S.~Zhang, M.~Wei, R.~Song, \href{https://www.sciencedirect.com/science/article/pii/S0925231219301316}{Robot skill acquisition in assembly process using deep reinforcement learning}, Neurocomputing 345 (2019) 92--102, deep Learning for Intelligent Sensing, Decision-Making and Control.
\newblock \href {https://doi.org/https://doi.org/10.1016/j.neucom.2019.01.087} {\path{doi:https://doi.org/10.1016/j.neucom.2019.01.087}}.
\newline\urlprefix\url{https://www.sciencedirect.com/science/article/pii/S0925231219301316}

\bibitem{ZHOU2025128669}
R.~Zhou, H.~Cao, J.~Huang, X.~Song, J.~Huang, Z.~Huang, \href{https://www.sciencedirect.com/science/article/pii/S0925231224014401}{Hybrid lane change strategy of autonomous vehicles based on soar cognitive architecture and deep reinforcement learning}, Neurocomputing 611 (2025) 128669.
\newblock \href {https://doi.org/https://doi.org/10.1016/j.neucom.2024.128669} {\path{doi:https://doi.org/10.1016/j.neucom.2024.128669}}.
\newline\urlprefix\url{https://www.sciencedirect.com/science/article/pii/S0925231224014401}

\bibitem{vinyals2019alpha}
O.~t. Vinyals, Grandmaster level in starcraft~ii using multi-agent reinforcement learning, Nature 575 (2019) 350--354.
\newblock \href {https://doi.org/10.1038/s41586-019-1724-z} {\path{doi:10.1038/s41586-019-1724-z}}.

\bibitem{YAN2022107688}
J.~Yan, Y.~Huang, A.~Gupta, A.~Gupta, C.~Liu, J.~Li, L.~Cheng, \href{https://www.sciencedirect.com/science/article/pii/S0045790622000106}{Energy-aware systems for real-time job scheduling in cloud data centers: A deep reinforcement learning approach}, Computers and Electrical Engineering 99 (2022) 107688.
\newblock \href {https://doi.org/https://doi.org/10.1016/j.compeleceng.2022.107688} {\path{doi:https://doi.org/10.1016/j.compeleceng.2022.107688}}.
\newline\urlprefix\url{https://www.sciencedirect.com/science/article/pii/S0045790622000106}

\bibitem{10.1162/neco.1997.9.8.1735}
S.~Hochreiter, J.~Schmidhuber, \href{https://doi.org/10.1162/neco.1997.9.8.1735}{Long short-term memory}, Neural Computation 9~(8) (1997) 1735--1780.
\newblock \href {http://arxiv.org/abs/https://direct.mit.edu/neco/article-pdf/9/8/1735/813796/neco.1997.9.8.1735.pdf} {\path{arXiv:https://direct.mit.edu/neco/article-pdf/9/8/1735/813796/neco.1997.9.8.1735.pdf}}, \href {https://doi.org/10.1162/neco.1997.9.8.1735} {\path{doi:10.1162/neco.1997.9.8.1735}}.
\newline\urlprefix\url{https://doi.org/10.1162/neco.1997.9.8.1735}

\bibitem{ELMAN1990179}
J.~L. Elman, \href{https://www.sciencedirect.com/science/article/pii/036402139090002E}{Finding structure in time}, Cognitive Science 14~(2) (1990) 179--211.
\newblock \href {https://doi.org/https://doi.org/10.1016/0364-0213(90)90002-E} {\path{doi:https://doi.org/10.1016/0364-0213(90)90002-E}}.
\newline\urlprefix\url{https://www.sciencedirect.com/science/article/pii/036402139090002E}

\bibitem{NIPS2017_3f5ee243}
A.~Vaswani, N.~Shazeer, N.~Parmar, J.~Uszkoreit, L.~Jones, A.~N. Gomez, L.~u. Kaiser, I.~Polosukhin, \href{https://proceedings.neurips.cc/paper_files/paper/2017/file/3f5ee243547dee91fbd053c1c4a845aa-Paper.pdf}{Attention is all you need}, in: I.~Guyon, U.~V. Luxburg, S.~Bengio, H.~Wallach, R.~Fergus, S.~Vishwanathan, R.~Garnett (Eds.), Advances in Neural Information Processing Systems, Vol.~30, Curran Associates, Inc., 2017.
\newline\urlprefix\url{https://proceedings.neurips.cc/paper_files/paper/2017/file/3f5ee243547dee91fbd053c1c4a845aa-Paper.pdf}

\bibitem{DBLP:conf/naacl/DevlinCLT19}
J.~Devlin, M.~Chang, K.~Lee, K.~Toutanova, \href{https://doi.org/10.18653/v1/n19-1423}{{BERT:} pre-training of deep bidirectional transformers for language understanding}, in: J.~Burstein, C.~Doran, T.~Solorio (Eds.), Proceedings of the 2019 Conference of the North American Chapter of the Association for Computational Linguistics: Human Language Technologies, {NAACL-HLT} 2019, Minneapolis, MN, USA, June 2-7, 2019, Volume 1 (Long and Short Papers), Association for Computational Linguistics, 2019, pp. 4171--4186.
\newblock \href {https://doi.org/10.18653/V1/N19-1423} {\path{doi:10.18653/V1/N19-1423}}.
\newline\urlprefix\url{https://doi.org/10.18653/v1/n19-1423}

\bibitem{Dosovitskiy2020AnII}
A.~Dosovitskiy, L.~Beyer, A.~Kolesnikov, D.~Weissenborn, X.~Zhai, T.~Unterthiner, M.~Dehghani, M.~Minderer, G.~Heigold, S.~Gelly, J.~Uszkoreit, N.~Houlsby, \href{https://api.semanticscholar.org/CorpusID:225039882}{An image is worth 16x16 words: Transformers for image recognition at scale}, ArXiv abs/2010.11929 (2020).
\newline\urlprefix\url{https://api.semanticscholar.org/CorpusID:225039882}

\bibitem{chen2021decisiontransformerreinforcementlearning}
L.~Chen, K.~Lu, A.~Rajeswaran, K.~Lee, A.~Grover, M.~Laskin, P.~Abbeel, A.~Srinivas, I.~Mordatch, \href{https://arxiv.org/abs/2106.01345}{Decision transformer: Reinforcement learning via sequence modeling} (2021).
\newblock \href {http://arxiv.org/abs/2106.01345} {\path{arXiv:2106.01345}}.
\newline\urlprefix\url{https://arxiv.org/abs/2106.01345}

\bibitem{neftci2019surrogategradientlearningspiking}
E.~O. Neftci, H.~Mostafa, F.~Zenke, \href{https://arxiv.org/abs/1901.09948}{Surrogate gradient learning in spiking neural networks} (2019).
\newblock \href {http://arxiv.org/abs/1901.09948} {\path{arXiv:1901.09948}}.
\newline\urlprefix\url{https://arxiv.org/abs/1901.09948}

\bibitem{8259423}
M.~Davies, N.~Srinivasa, T.-H. Lin, G.~Chinya, Y.~Cao, S.~H. Choday, G.~Dimou, P.~Joshi, N.~Imam, S.~Jain, Y.~Liao, C.-K. Lin, A.~Lines, R.~Liu, D.~Mathaikutty, S.~McCoy, A.~Paul, J.~Tse, G.~Venkataramanan, Y.-H. Weng, A.~Wild, Y.~Yang, H.~Wang, Loihi: A neuromorphic manycore processor with on-chip learning, IEEE Micro 38~(1) (2018) 82--99.
\newblock \href {https://doi.org/10.1109/MM.2018.112130359} {\path{doi:10.1109/MM.2018.112130359}}.

\bibitem{doi:10.1126/science.1254642}
P.~A. Merolla, J.~V. Arthur, R.~Alvarez-Icaza, A.~S. Cassidy, J.~Sawada, F.~Akopyan, B.~L. Jackson, N.~Imam, C.~Guo, Y.~Nakamura, B.~Brezzo, I.~Vo, S.~K. Esser, R.~Appuswamy, B.~Taba, A.~Amir, M.~D. Flickner, W.~P. Risk, R.~Manohar, D.~S. Modha, \href{https://www.science.org/doi/abs/10.1126/science.1254642}{A million spiking-neuron integrated circuit with a scalable communication network and interface}, Science 345~(6197) (2014) 668--673.
\newblock \href {http://arxiv.org/abs/https://www.science.org/doi/pdf/10.1126/science.1254642} {\path{arXiv:https://www.science.org/doi/pdf/10.1126/science.1254642}}, \href {https://doi.org/10.1126/science.1254642} {\path{doi:10.1126/science.1254642}}.
\newline\urlprefix\url{https://www.science.org/doi/abs/10.1126/science.1254642}

\bibitem{10.1145/3320288.3320304}
P.~Blouw, X.~Choo, E.~Hunsberger, C.~Eliasmith, \href{https://doi.org/10.1145/3320288.3320304}{Benchmarking keyword spotting efficiency on neuromorphic hardware}, in: Proceedings of the 7th Annual Neuro-Inspired Computational Elements Workshop, NICE '19, Association for Computing Machinery, New York, NY, USA, 2019.
\newblock \href {https://doi.org/10.1145/3320288.3320304} {\path{doi:10.1145/3320288.3320304}}.
\newline\urlprefix\url{https://doi.org/10.1145/3320288.3320304}

\bibitem{doi:10.1073/pnas.1604850113}
S.~K. Esser, P.~A. Merolla, J.~V. Arthur, A.~S. Cassidy, R.~Appuswamy, A.~Andreopoulos, D.~J. Berg, J.~L. McKinstry, T.~Melano, D.~R. Barch, C.~di~Nolfo, P.~Datta, A.~Amir, B.~Taba, M.~D. Flickner, D.~S. Modha, \href{https://www.pnas.org/doi/abs/10.1073/pnas.1604850113}{Convolutional networks for fast, energy-efficient neuromorphic computing}, Proceedings of the National Academy of Sciences 113~(41) (2016) 11441--11446.
\newblock \href {http://arxiv.org/abs/https://www.pnas.org/doi/pdf/10.1073/pnas.1604850113} {\path{arXiv:https://www.pnas.org/doi/pdf/10.1073/pnas.1604850113}}, \href {https://doi.org/10.1073/pnas.1604850113} {\path{doi:10.1073/pnas.1604850113}}.
\newline\urlprefix\url{https://www.pnas.org/doi/abs/10.1073/pnas.1604850113}

\bibitem{10.3389/fnins.2022.1018006}
A.~Rostami, B.~Vogginger, Y.~Yan, C.~G. Mayr, \href{https://www.frontiersin.org/journals/neuroscience/articles/10.3389/fnins.2022.1018006}{E-prop on spinnaker 2: Exploring online learning in spiking rnns on neuromorphic hardware}, Frontiers in Neuroscience Volume 16 - 2022 (2022).
\newblock \href {https://doi.org/10.3389/fnins.2022.1018006} {\path{doi:10.3389/fnins.2022.1018006}}.
\newline\urlprefix\url{https://www.frontiersin.org/journals/neuroscience/articles/10.3389/fnins.2022.1018006}

\bibitem{10.3389/fnins.2022.877701}
G.~Wu, D.~Liang, S.~Luan, J.~Wang, \href{https://www.frontiersin.org/journals/neuroscience/articles/10.3389/fnins.2022.877701}{Training spiking neural networks for reinforcement learning tasks with temporal coding method}, Frontiers in Neuroscience Volume 16 - 2022 (2022).
\newblock \href {https://doi.org/10.3389/fnins.2022.877701} {\path{doi:10.3389/fnins.2022.877701}}.
\newline\urlprefix\url{https://www.frontiersin.org/journals/neuroscience/articles/10.3389/fnins.2022.877701}

\bibitem{tang2020reinforcementcolearningdeepspiking}
G.~Tang, N.~Kumar, K.~P. Michmizos, \href{https://arxiv.org/abs/2003.01157}{Reinforcement co-learning of deep and spiking neural networks for energy-efficient mapless navigation with neuromorphic hardware} (2020).
\newblock \href {http://arxiv.org/abs/2003.01157} {\path{arXiv:2003.01157}}.
\newline\urlprefix\url{https://arxiv.org/abs/2003.01157}

\bibitem{shrestha2018slayerspikelayererror}
S.~B. Shrestha, G.~Orchard, \href{https://arxiv.org/abs/1810.08646}{Slayer: Spike layer error reassignment in time} (2018).
\newblock \href {http://arxiv.org/abs/1810.08646} {\path{arXiv:1810.08646}}.
\newline\urlprefix\url{https://arxiv.org/abs/1810.08646}

\bibitem{bellec2020solution}
G.~Bellec, F.~Scherr, A.~Subramoney, E.~Hajek, D.~Salaj, R.~Legenstein, W.~Maass, A solution to the learning dilemma for recurrent networks of spiking neurons, Nature Communications 11 (07 2020).
\newblock \href {https://doi.org/10.1038/s41467-020-17236-y} {\path{doi:10.1038/s41467-020-17236-y}}.

\bibitem{10.3389/fnins.2018.00331}
Y.~Wu, L.~Deng, G.~Li, J.~Zhu, L.~Shi, \href{https://www.frontiersin.org/journals/neuroscience/articles/10.3389/fnins.2018.00331}{Spatio-temporal backpropagation for training high-performance spiking neural networks}, Frontiers in Neuroscience Volume 12 - 2018 (2018).
\newblock \href {https://doi.org/10.3389/fnins.2018.00331} {\path{doi:10.3389/fnins.2018.00331}}.
\newline\urlprefix\url{https://www.frontiersin.org/journals/neuroscience/articles/10.3389/fnins.2018.00331}

\bibitem{fang2021incorporatinglearnablemembranetime}
W.~Fang, Z.~Yu, Y.~Chen, T.~Masquelier, T.~Huang, Y.~Tian, \href{https://arxiv.org/abs/2007.05785}{Incorporating learnable membrane time constant to enhance learning of spiking neural networks} (2021).
\newblock \href {http://arxiv.org/abs/2007.05785} {\path{arXiv:2007.05785}}.
\newline\urlprefix\url{https://arxiv.org/abs/2007.05785}

\bibitem{GAO2024128268}
S.~Gao, X.~Fan, X.~Deng, Z.~Hong, H.~Zhou, Z.~Zhu, \href{https://www.sciencedirect.com/science/article/pii/S0925231224010397}{Te-spikformer:temporal-enhanced spiking neural network with transformer}, Neurocomputing 602 (2024) 128268.
\newblock \href {https://doi.org/https://doi.org/10.1016/j.neucom.2024.128268} {\path{doi:https://doi.org/10.1016/j.neucom.2024.128268}}.
\newline\urlprefix\url{https://www.sciencedirect.com/science/article/pii/S0925231224010397}

\bibitem{gerstner2002spiking}
W.~Gerstner, W.~M. Kistler, \href{https://psycnet.apa.org/doi/10.1017/CBO9780511815706}{Spiking Neuron Models: Single Neurons, Populations, Plasticity}, Cambridge University Press, 2002.
\newline\urlprefix\url{https://psycnet.apa.org/doi/10.1017/CBO9780511815706}

\bibitem{janner2021offlinereinforcementlearningbig}
M.~Janner, Q.~Li, S.~Levine, \href{https://arxiv.org/abs/2106.02039}{Offline reinforcement learning as one big sequence modeling problem} (2021).
\newblock \href {http://arxiv.org/abs/2106.02039} {\path{arXiv:2106.02039}}.
\newline\urlprefix\url{https://arxiv.org/abs/2106.02039}

\bibitem{Ghanem2023}
A.~Ghanem, P.~Ciblat, M.~Ghogho, Multi-objective decision transformers for offline reinforcement learning (08 2023).
\newblock \href {https://doi.org/10.48550/arXiv.2308.16379} {\path{doi:10.48550/arXiv.2308.16379}}.

\bibitem{Correia2023}
A.~Correia, L.~Alexandre, Hierarchical decision transformer, 2023, pp. 1661--1666.
\newblock \href {https://doi.org/10.1109/IROS55552.2023.10342230} {\path{doi:10.1109/IROS55552.2023.10342230}}.

\bibitem{pmlr-v205-konan23a}
S.~G. Konan, E.~Seraj, M.~Gombolay, \href{https://proceedings.mlr.press/v205/konan23a.html}{Contrastive decision transformers}, in: K.~Liu, D.~Kulic, J.~Ichnowski (Eds.), Proceedings of The 6th Conference on Robot Learning, Vol. 205 of Proceedings of Machine Learning Research, PMLR, 2023, pp. 2159--2169.
\newline\urlprefix\url{https://proceedings.mlr.press/v205/konan23a.html}

\bibitem{li2024uncertaintyawaredecisiontransformerstochastic}
Z.~Li, F.~Nie, Q.~Sun, F.~Da, H.~Zhao, \href{https://arxiv.org/abs/2309.16397}{Uncertainty-aware decision transformer for stochastic driving environments} (2024).
\newblock \href {http://arxiv.org/abs/2309.16397} {\path{arXiv:2309.16397}}.
\newline\urlprefix\url{https://arxiv.org/abs/2309.16397}

\bibitem{brohan2023rt1roboticstransformerrealworld}
A.~Brohan, N.~Brown, J.~Carbajal, Y.~Chebotar, J.~Dabis, C.~Finn, K.~Gopalakrishnan, K.~Hausman, A.~Herzog, J.~Hsu, J.~Ibarz, B.~Ichter, A.~Irpan, T.~Jackson, S.~Jesmonth, N.~J. Joshi, R.~Julian, D.~Kalashnikov, Y.~Kuang, I.~Leal, K.-H. Lee, S.~Levine, Y.~Lu, U.~Malla, D.~Manjunath, I.~Mordatch, O.~Nachum, C.~Parada, J.~Peralta, E.~Perez, K.~Pertsch, J.~Quiambao, K.~Rao, M.~Ryoo, G.~Salazar, P.~Sanketi, K.~Sayed, J.~Singh, S.~Sontakke, A.~Stone, C.~Tan, H.~Tran, V.~Vanhoucke, S.~Vega, Q.~Vuong, F.~Xia, T.~Xiao, P.~Xu, S.~Xu, T.~Yu, B.~Zitkovich, \href{https://arxiv.org/abs/2212.06817}{Rt-1: Robotics transformer for real-world control at scale} (2023).
\newblock \href {http://arxiv.org/abs/2212.06817} {\path{arXiv:2212.06817}}.
\newline\urlprefix\url{https://arxiv.org/abs/2212.06817}

\bibitem{Mnih2015HumanlevelCT}
V.~Mnih, K.~Kavukcuoglu, D.~Silver, A.~A. Rusu, J.~Veness, M.~G. Bellemare, A.~Graves, M.~A. Riedmiller, A.~K. Fidjeland, G.~Ostrovski, S.~Petersen, C.~Beattie, A.~Sadik, I.~Antonoglou, H.~King, D.~Kumaran, D.~Wierstra, S.~Legg, D.~Hassabis, \href{https://api.semanticscholar.org/CorpusID:205242740}{Human-level control through deep reinforcement learning}, Nature 518 (2015) 529--533.
\newline\urlprefix\url{https://api.semanticscholar.org/CorpusID:205242740}

\bibitem{712192}
R.~Sutton, A.~Barto, Reinforcement learning: An introduction, IEEE Transactions on Neural Networks 9~(5) (1998) 1054--1054.
\newblock \href {https://doi.org/10.1109/TNN.1998.712192} {\path{doi:10.1109/TNN.1998.712192}}.

\bibitem{Roy2019TowardsSM}
K.~Roy, A.~R. Jaiswal, P.~Panda, \href{https://api.semanticscholar.org/CorpusID:208329736}{Towards spike-based machine intelligence with neuromorphic computing}, Nature 575 (2019) 607 -- 617.
\newline\urlprefix\url{https://api.semanticscholar.org/CorpusID:208329736}

\bibitem{li2022spikeformernovelarchitecturetraining}
Y.~Li, Y.~Lei, X.~Yang, \href{https://arxiv.org/abs/2211.10686}{Spikeformer: A novel architecture for training high-performance low-latency spiking neural network} (2022).
\newblock \href {http://arxiv.org/abs/2211.10686} {\path{arXiv:2211.10686}}.
\newline\urlprefix\url{https://arxiv.org/abs/2211.10686}

\bibitem{zhou2022spikformerspikingneuralnetwork}
Z.~Zhou, Y.~Zhu, C.~He, Y.~Wang, S.~Yan, Y.~Tian, L.~Yuan, \href{https://arxiv.org/abs/2209.15425}{Spikformer: When spiking neural network meets transformer} (2022).
\newblock \href {http://arxiv.org/abs/2209.15425} {\path{arXiv:2209.15425}}.
\newline\urlprefix\url{https://arxiv.org/abs/2209.15425}

\bibitem{fu2021d4rldatasetsdeepdatadriven}
J.~Fu, A.~Kumar, O.~Nachum, G.~Tucker, S.~Levine, \href{https://arxiv.org/abs/2004.07219}{D4rl: Datasets for deep data-driven reinforcement learning} (2021).
\newblock \href {http://arxiv.org/abs/2004.07219} {\path{arXiv:2004.07219}}.
\newline\urlprefix\url{https://arxiv.org/abs/2004.07219}

\bibitem{4082128}
P.~E. Hart, N.~J. Nilsson, B.~Raphael, A formal basis for the heuristic determination of minimum cost paths, IEEE Transactions on Systems Science and Cybernetics 4~(2) (1968) 100--107.
\newblock \href {https://doi.org/10.1109/TSSC.1968.300136} {\path{doi:10.1109/TSSC.1968.300136}}.

\bibitem{loshchilov2019decoupledweightdecayregularization}
I.~Loshchilov, F.~Hutter, \href{https://arxiv.org/abs/1711.05101}{Decoupled weight decay regularization} (2019).
\newblock \href {http://arxiv.org/abs/1711.05101} {\path{arXiv:1711.05101}}.
\newline\urlprefix\url{https://arxiv.org/abs/1711.05101}

\bibitem{loshchilov2017sgdrstochasticgradientdescent}
I.~Loshchilov, F.~Hutter, \href{https://arxiv.org/abs/1608.03983}{Sgdr: Stochastic gradient descent with warm restarts} (2017).
\newblock \href {http://arxiv.org/abs/1608.03983} {\path{arXiv:1608.03983}}.
\newline\urlprefix\url{https://arxiv.org/abs/1608.03983}

\end{thebibliography}

\end{document}